\documentclass[sigconf]{acmart}

\AtBeginDocument{%
  \providecommand\BibTeX{{%
    \normalfont B\kern-0.5em{\scshape i\kern-0.25em b}\kern-0.8em\TeX}}}

\usepackage[normalem]{ulem}
\usepackage{booktabs} 
\usepackage{subcaption}

\usepackage[english]{babel}
\usepackage[utf8x]{inputenc}
\usepackage[T1]{fontenc}
\usepackage{comment} 

\usepackage{amsmath}
\usepackage{graphicx}
\usepackage{xcolor}
\usepackage[colorinlistoftodos]{todonotes}
\usepackage{multirow}
\usepackage{array} 
\usepackage{mwe}
\usepackage{xcolor, soul}
\sethlcolor{red}

\newcommand{\el}{et al.}

\copyrightyear{2023}
\acmYear{2023}
\setcopyright{acmlicensed}\acmConference[GECCO '23]{Genetic and Evolutionary Computation Conference}{July 15--19, 2023}{Lisbon, Portugal}
\acmBooktitle{Genetic and Evolutionary Computation Conference (GECCO '23), July 15--19, 2023, Lisbon, Portugal}
\acmPrice{15.00}
\acmDOI{10.1145/3583131.3590352}
\acmISBN{979-8-4007-0119-1/23/07}

\begin{document}

\title[Evolving Flying Machines in Minecraft Using QD]{Evolving Flying Machines in Minecraft Using Quality Diversity}

\author{Alejandro Medina, Melanie Richey, Mark Mueller, and Jacob Schrum}
\orcid{0009-0005-7618-4947,0009-0007-1617-838X,0009-0000-5464-4723,0000-0002-7315-0515}
\affiliation{%
  \institution{Southwestern University}
  \streetaddress{1001 E. University Ave}
  \city{Georgetown} 
  \state{Texas}
  \country{USA}
  \postcode{78626}
}
\email{{medina4,richey2,muellerm,schrum2}@southwestern.edu}


\begin{abstract}
\emph{Minecraft} is a great testbed for human creativity that has inspired the design of various structures and even functioning machines, including flying machines. \emph{EvoCraft} is an API for programmatically generating structures in \emph{Minecraft}, but the initial work in this domain was not capable of evolving flying machines. This paper applies fitness-based evolution and quality diversity search in order to evolve flying machines. Although fitness alone can occasionally produce flying machines, thanks in part to a more sophisticated fitness function than was used previously, the quality diversity algorithm MAP-Elites is capable of discovering flying machines much more reliably, at least when an appropriate behavior characterization is used to guide the search for diverse solutions.
\end{abstract}

%
%



\maketitle

\section{Introduction}


The game \emph{Minecraft} offers a rich world that allows players to apply their creativity. 
The aesthetic appeal of human creations is one driver of interest in the game, yet machines with interesting functional properties can also be created. Incredibly complex machines have been created by humans, such as a walking robot\footnote{\url{https://youtu.be/GPbE6fnNfSA}} and a functioning word processor,\footnote{\url{https://youtu.be/g_ULtNYRCbg}} but a more modest, yet still impressive, human designed structure is a flying machine: a machine that perpetually moves in a given direction. Such machines can be of practical use to players when navigating the game world, though the main reason they are of interest to researchers is that their design is a non-trivial engineering problem for Artificial Intelligence to solve.

Creations like these inspired the design of the \emph{EvoCraft} API \cite{grbic2021evocraft}, which makes it possible to programmatically place blocks in the game. The work introducing this API demonstrated that it could be used to evolve shapes satisfying simple objectives, such as spanning space to connect to a target block, yet they also tried and failed to evolve flying machines.
Thus, the challenge of automatically generating flying machines remained open, until now.

This paper uses a different shape encoding and a more sophisticated fitness function to demonstrate that evolving flying machines is possible, but in order to reliably produce flying machines, the quality diversity algorithm MAP-Elites \cite{mouret:arxiv15} is needed. MAP-Elites maintains a diverse archive of solutions throughout evolution, and when this archive is structured appropriately, a single experimental run can create various machines that fly in different directions. However, the results also demonstrate the importance of choosing the right behavior characterization for MAP-Elites, since the wrong characterization can be worse than or no better than pure fitness.

The achievement of evolving flying machines is a first step toward more open-ended evolution of a wider variety of interesting functional and aesthetic artefacts. \emph{Minecraft} players have a long history of impressing human viewers with the fruits of their labor, but now it is time to show off the range of creative expression that evolutionary computation is capable of.



\section{Related Work}
\label{sec:related}

This research builds on previous work in procedural content generation and on applications of quality diversity algorithms.

\subsection{Procedural Content Generation}
\label{sec:pcg}



Procedural content generation (PCG) has been used to generate a wide variety of game content \cite{shaker2016procedural}. 
Levels have been generated for many games. A popular choice is \emph{Super Mario Bros}, where long short-term memory networks \cite{summerville2016super}, generative adversarial networks (GANs) \cite{volz:gecco2018}, grammatical evolution \cite{shaker2012evolving}, and more have been applied. 
Other types of content can also be generated, such as the flowers bred for their aesthetic qualities in the social game \emph{Petalz} \cite{risi:tciaig16petalz}.
However, users can influence the evolution of both the aesthetic and functional properties of game content, as done with the weapon firing patterns in the \emph{Galactic Arms Race} video game \cite{hastings2009automatic}.


PCG has also been used in \emph{Minecraft}. The WorldGAN approach of Awiszus \el~\cite{awiszus:cog2021} can generate new 3D world segments based on a single training input. Sudhakaran \el~used neural cellular automata (NCAs) to regenerate learned structures from starting cells, including functional machines and 3D artefacts~\cite{Sudhakaran:alife2021}. This NCA approach only regenerated human designs, and did not discover them from scratch. Still, the range of artefacts generated, e.g.\ castles, temples, trees, buildings, and flying machines, was impressive. 

\subsection{Quality Diversity}
\label{sec:qd}



An example of original shapes being evolved in \emph{Minecraft} comes from work by Barthet \el~\cite{Barthet:togEA} where compositional pattern producing networks (CPPNs \cite{stanley:gpem2007}) generate static structures evolved through novelty search \cite{lehman:ecj2011}. The novelty is assessed via the latent representation of the shape derived from an autoencoder. Combining novelty search with a mechanism (the autoencoder) to focus the search on the most relevant novel shapes connects this approach to a broader search philosophy known as quality diversity algorithms.

Quality diversity (QD) algorithms, also known as illumination algorithms \cite{mouret:arxiv15}, are used to find good solutions to difficult problems, but value different types of good solutions. QD has been used to generate decks for the game \emph{HearthStone} \cite{fontaine:gecco20}, levels for \emph{Super Mario Bros.}~and \emph{The Legend of Zelda} \cite{schrum:gecco2020cppn2gan}, 
multi-legged robot gaits \cite{cully:nature15},
and more \cite{gravina:cog2019pcgqdsurvey}. QD searches for a wide array of solutions with different behaviors, where each solution is the best in its particular niche. 


This paper uses MAP-Elites (Multi-dimensional Archive of Phenotypic Elites \cite{mouret:arxiv15}) since it stores various solutions in an organized archive. There are multiple variations of MAP-Elites including CMA-ME \cite{fontaine2020illuminating}, Multi-Emitter MAP-Elites \cite{cully:gecco2020}, and DQD \cite{tjanaka:gecco2022}, which enhance the search process in various ways, but vanilla MAP-Elites proved sufficient to produce a variety of flying machines. 

\section{Minecraft}
\label{sec:minecraft}

\begin{table}[t]
\caption{\label{tab:blocktypes} Blocks Used in Flying Machines}
The observer is only available in the observer block set. All other blocks are available in both the original and observer block sets.
\begin{tabular}{|l|c|p{0.53\columnwidth}|}
\hline
Name & Image & Description \\ \hline

Redstone Block & 
\raisebox{-.6\height}{\includegraphics[width=10mm]{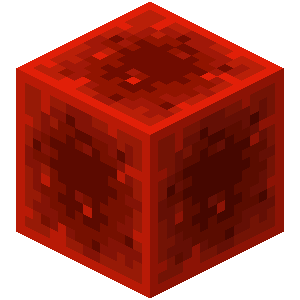}}
& Provides power to adjacent blocks \\ \hline

Slime Block & 
\raisebox{-.6\height}{\includegraphics[width=10mm]{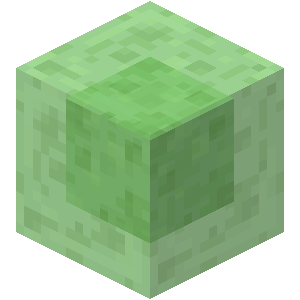}}
& Sticks to other blocks causing them to move together \\ \hline

Quartz Block & 
\raisebox{-.6\height}{\includegraphics[width=10mm]{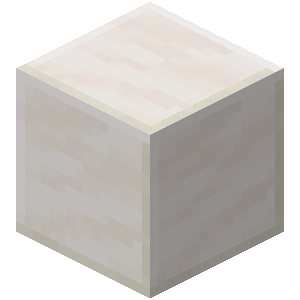}}
& Provides structural support \\ \hline

Piston & 
\raisebox{-.6\height}{\includegraphics[width=10mm]{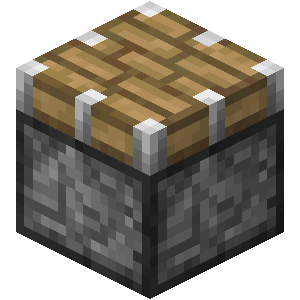}}
& Pushes whatever is in front of it one block distance when activated with a power source \\ \hline

Sticky Piston  & 
\raisebox{-.6\height}{\includegraphics[width=10mm]{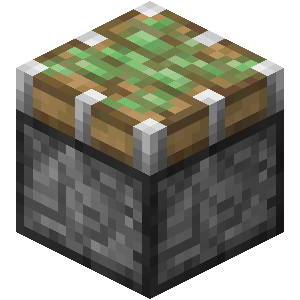}}
& A piston that also pulls attached blocks back whenever it loses power \\ \hline

Observer  & 
\raisebox{-.6\height}{\includegraphics[width=10mm]{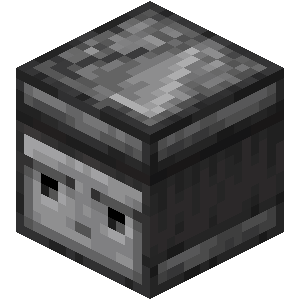}}
& Detects movement at one end and supplies power in response at the opposite end \\ \hline

\end{tabular}
\end{table}


\emph{Minecraft}\footnote{\url{https://www.minecraft.net}} is a popular sandbox video game known for allowing users to build whatever they can imagine. Various competitions have emerged to test the limits of AI techniques applied to the game. The \emph{Generative Design in Minecraft Challenge}\footnote{\url{https://gendesignmc.engineering.nyu.edu/}} focuses on generating settlements that resemble human-designed cities. The \emph{MineRL Diamond Competition}\footnote{\url{https://minerl.io/diamond/}} focuses on training agents to find diamonds. This paper is inspired by the \emph{Evocraft Open-Endedness Challenge},\footnote{\url{https://evocraft.life/}} which focuses on continuously generating interesting and novel artefacts. The  paper \cite{grbic2021evocraft} that introduced the API used in the challenge demonstrates examples of successful evolution in \emph{Minecraft}, but was unable to evolve flying machines. Note that most \emph{Minecraft} blocks do not experience gravity,
so a ``flying'' machine is really just any collection of blocks that keeps moving in the same direction forever.
A flying machine is complex enough that even a human would struggle to make one without previous knowledge, but because flying machines consist of only a handful of blocks, they represent a reasonable challenge for evolution to overcome.



In the original \emph{EvoCraft} paper \cite{grbic2021evocraft}, the attempt to evolve flying machines used a limited block set: redstone block, quartz block, slime block, piston, and sticky piston.
Throughout the paper, this collection of blocks is referred to as the \emph{original} block set.
Images and descriptions of each of these blocks are in Table \ref{tab:blocktypes}.

For a machine to fly, pistons must move in a particular order to push the shape forward. At least two pistons must face opposite directions, and certain blocks need to stick to each other so that other parts of the structure are pushed or pulled along. 
If the pistons and power sources are not in the right positions, their timing will not be synchronized, and the shape will not move.


When humans create flying machines, observer blocks are often used as well (also in Table \ref{tab:blocktypes}). 
An observer has a sensor at one end that detects movement, and produces a redstone charge at the other end in response to movement. 
The use of observers makes it slightly easier to create flying machines, and also leads to machines that can behave slightly differently. The observer was added to the original block set to create a set known as the \emph{observer} block set. Both sets are compared in experiments described in Section \ref{sec:experiments}. Another block from \emph{Minecraft} whose inclusion was considered is the honey block, but despite some small differences, honey behaves nearly identically to slime in the context of flying machine creation, so honey was excluded. Its inclusion in future work could be interesting.


\section{Approach}
\label{sec:approach}

This section covers the genome encoding used for generating shapes, the fitness function used for evolving shapes, and how MAP-Elites helped generate flying machines.

\subsection{Genome Encoding}
\label{sec:genome}


For the flying machine task, the original work in \emph{EvoCraft} \cite{grbic2021evocraft} evolved neural networks to generate the evaluated shapes. The networks were queried in a manner similar to compositional pattern producing networks (CPPNs~\cite{stanley:gpem2007}), in that block locations were given as inputs, and the real-valued outputs specified the presence, block type, and orientation of the block. Our preliminary experiments applied CPPNs, but we ultimately opted for a simpler encoding that directly evolves real-valued vectors rather than have a neural network produce these values.

\begin{figure}[t]
\centering
\includegraphics[width=0.9\columnwidth]{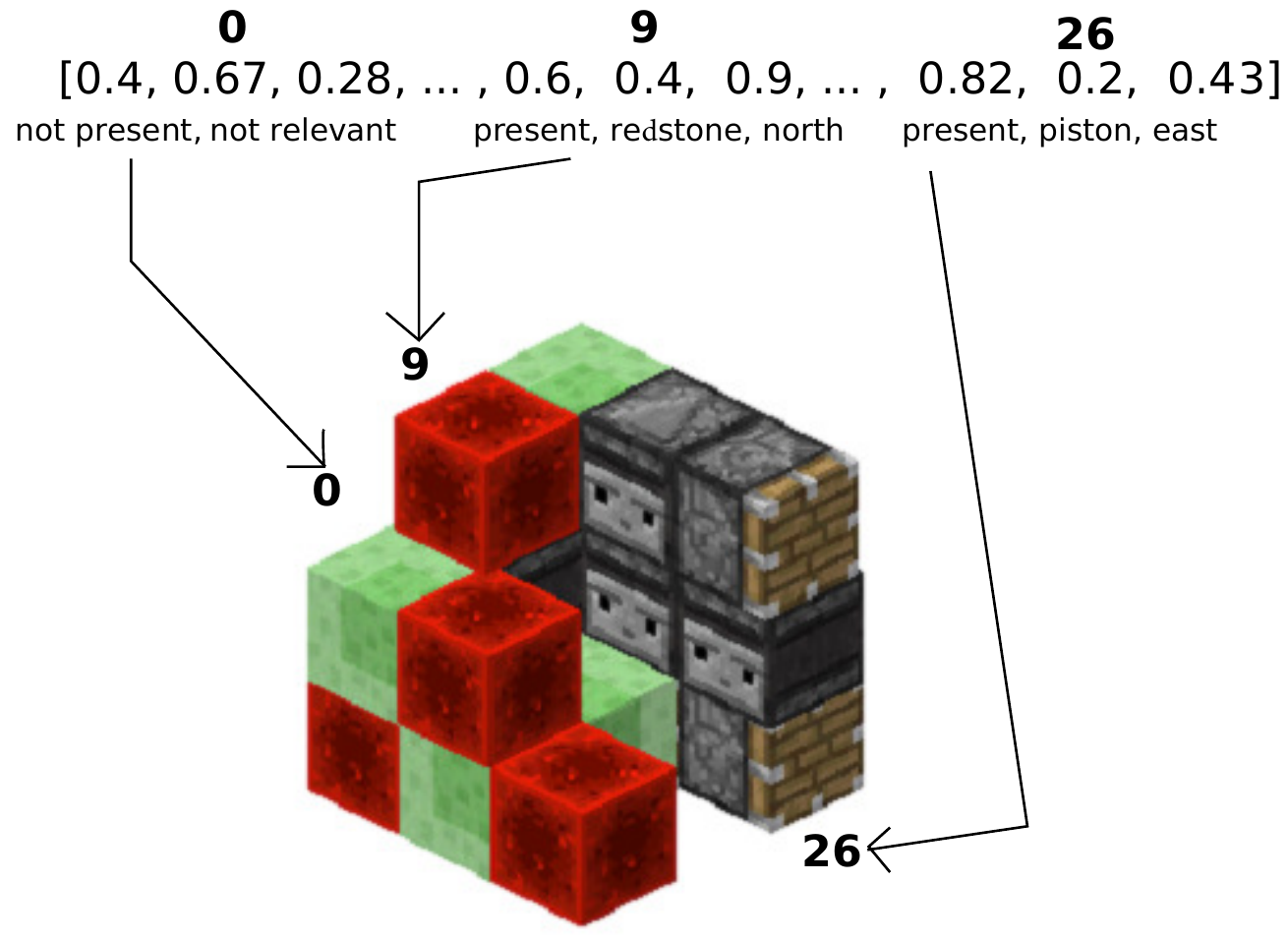}
\caption{Genome Encoding: three sections of a vector are shown
encoding three blocks in a larger shape. Each set of three values
corresponds to one position in the shape. Block 0 has a presence
value of 0.4, so it is not in the shape at all and the next two
numbers are ignored. Blocks 9 and 26 have presence values above
0.5, so the following numbers determine block type and orientation.}
\label{fig:genomeEncoding}
\end{figure}

Each genome is a real-valued vector of three values per block of the shape, each in the range [0,1] (Fig.~\ref{fig:genomeEncoding}). Each group of three values corresponds to a block at a pre-determined position based on location in the genome. The three values represent the block presence, block type, and block orientation. Block presence values above 0.5 mean a block is present at the particular location of the shape. For block type, the range of possible values is evenly divided among the number of possible block types in whichever block set is being used, so that each interval corresponds to a different block type. Similarly, the range of values for block orientation is evenly divided among the six possibilities: up, down, north, south, east, and west. A discrete encoding could have been used instead. This real-valued encoding was used primarily for the sake of comparison with CPPNs, even though such comparisons are not included in this paper. Use of a discrete encoding is an option for future work.

Block orientation is actually irrelevant for most blocks, since e.g.\ redstone blocks do not behave differently in different orientations. However, when generating flying machines, orientation is vital for the blocks where it has an effect: pistons, sticky pistons, and observers.
The way a piston or sticky piston faces determines where it will push other blocks, and the orientation of observers determines where they sense movement from, and where they direct a redstone charge when movement is sensed.


Unfortunately, the \emph{Minecraft} mod that \emph{EvoCraft} is built on top of has a bug that prevents observer blocks from being placed in the world with an up or down orientation. Instead, both of these options result in the block facing north.
This bug was not noticed until extensive experiments were well underway. Although evolution finds ways around this limitation, it does restrict the range of flying machines that are possible to generate.

\subsection{Fitness Function}
\label{sec:fitness}



The fitness function is based off the 
original work in \emph{EvoCraft} \cite{grbic2021evocraft}, in which the center of mass of a shape was evaluated after 10 seconds to measure how far the shape had moved. However, confirming what was discovered in the original \emph{EvoCraft} work, this fitness function was not successful in preliminary experiments.

Successful flying machines do not simply move in one direction, but rather exhibit oscillating movement that nonetheless gradually pushes the shape in one direction. Therefore, the modified fitness function used in this paper
evaluates the center of mass every second, and accumulates the changes over time.
However, this periodic evaluation does not create a fitness gradient toward flying machines either, since simple oscillating movement can accrue great fitness without the machine ever actually flying away. Rewarding oscillation does at least make stumbling across flying behavior more likely, but to actually reward flying when it happens, the fitness function has a special case that notices when at least 6 blocks of the shape have left the region surrounding where the shape initially spawned. The resulting fitness score exceeds what the shape could ever earn through oscillation alone, but assures that actual flying machines are not incorrectly discarded in favor of shapes that simply oscillate a lot. One final tweak to the fitness function is a small penalty for any blocks that are left behind after the shape has flown away, which discourages superfluous blocks within the shape. Formally, the fitness $F$ of a shape $s$ is:



\begin{equation}
    F(s)=
    \begin{cases}
        F_{\max} - 0.1 \times \text{count}(s) & \text{if over 6 blocks leave}\\
        \Sigma_{t=1}^n \text{dist}(\Vec{p}_{t-1}(s),\Vec{p}_{t}(s)) & \text{otherwise}
    \end{cases}
\end{equation}
$F_{\max}$ is the maximum fitness awarded for flying and $\text{count}(s)$ is the number of blocks remaining in the evaluation area after the other blocks fly away. Subtraction of this term penalizes a shape for leaving blocks behind at a cost of 0.1 per block. If the shape does not fly away, then $\Vec{p}_{t}(s)$ is the 3D vector center of mass of the shape $s$ at time $t$, and $\text{dist}(\cdot,\cdot)$ is the Euclidean distance between two points, so the standard fitness is the accumulation of distance measurements between $n$ periodic queries on the center of mass.

Once evaluation is complete, a large area around the shape is cleared. This area exceeds the boundaries used to check when a flying machine has flown away. Clearing such a large area removes the shape if it succeeded at flying away, or at least assures that no left over blocks interfere with evaluation of the next shape.

The periodic polling of a shape's center of mass makes fitness evaluation nondeterministic and thus noisy, but if a machine succeeds at flying, then this will occur reliably. The exact amount of movement measured between each poll of the center of mass may fluctuate, but the special case for detecting when a machine has flown away is not noisy at all.

The lineage of a successful flying machine will thus typically contain three stages:
stagnant, oscillating, then flying. No movement is displayed by stagnant machines, or there could be some small movement at initialization followed by stillness.
For the sake of efficiency, evaluation terminates early if there is no change in the center of mass between subsequent readings.
An oscillating machine accumulates fitness until the end of evaluation, which is fixed at a maximum of 10 seconds, as in the original \emph{EvoCraft} paper \cite{grbic2021evocraft}. A flying machine does not require the full 10 seconds for evaluation, since it leaves the evaluation area before time elapses, though there can be some variation in how long a machine takes to fly away. However, flying machines are not rated in terms of speed. The only fitness distinction between flying machines is the number of blocks they leave behind.

As mentioned above, rewarding oscillation can help a machine stumble across the ability to fly, but 
sustained movement in one direction
is not explicitly promoted by the default case of the fitness function. Also, a population already containing flying machines does not have any direction for improvement beyond eliminating superfluous blocks, and is thus encouraged to converge around the first solution discovered by a pure fitness-based approach. However, a quality diversity approach can work around these limitations.


\subsection{MAP-Elites}
\label{sec:mapelites}



The Multi-dimensional Archive of Phenotypic Elites (MAP-Elites \cite{mouret:arxiv15}) does not focus on just a singular objective. Rather,
the most fit individuals 
in various niches are kept, resulting in a variety of quality artefacts. Such an approach exerts productive selection pressure even when there is a stagnant fitness landscape.

The archive is structured according to several orthogonal dimensions associated with a \emph{behavior characterization}, which is a means of quantitatively assessing certain features of a candidate solution. Typically, a behavior characterization produces a vector, and each score within the vector corresponds to one dimension of the multi-dimensional archive. Each dimension is divided into several intervals, so that every bin stores one artefact with a unique combination of features. Only the most fit individual with a given set of features can be kept in each bin.


Initially, 100 random samples are generated and categorized into the bins within the archive based on their features. Since each bin can only hold one individual, new candidates for an occupied bin either replace the previous occupant or are discarded based on who is fitter. 
Some bins may not be filled at initialization, and some may be impossible to fill. After initialization, new artefacts are uniformly sampled from the bins
to create offspring. Standard MAP-Elites does not allow crossover, but the version in this paper probabilistically creates offspring from sampling two random bins and mating the elites to create a child, which is then randomly mutated. Otherwise, each offspring is simply a mutated clone of a single parent. Either way, the resulting offspring is categorized into a bin based on its behavior characterization, and will either fill an empty bin, replace the current occupant, or be discarded as during initialization.


In general, this algorithm has the potential to productively search for solutions even when a fitness function would get stuck or lead the search astray, but the success of MAP-Elites depends on various details, as described in the experiments below.

\section{Experiments}
\label{sec:experiments}


This section covers the behavior characterizations used by MAP-Elites, how these are compared to a pure fitness-based approach, and various parameter settings. Code for running all experiments is included within the MM-NEAT repository available online at \url{https://github.com/schrum2/MM-NEAT}.

\subsection{Behavior Characterizations}


Although the fitness function defined in Section~\ref{sec:fitness} can recognize when a shape is flying, it does not create a smooth fitness gradient from stationary to flying. It rewards oscillation, but offspring still jump rather suddenly and unpredictably from oscillating to flying. To help discover shapes that make this jump, it is important to explore a diverse variety of shapes, which is the intent of each of the behavior characterizations described below. 

\begin{enumerate}

\item {\bf Block Count:}
This is the simplest behavior characterization. The blocks that make up a shape are counted, and there is one bin in the archive for each count of blocks. Referred to as {\tt ME.C} in results.

\item {\bf Block Count and Negative Space:}
The negative space within a shape is the number of air blocks that are confined by solid blocks in the shape. 
Assuming shapes are generated within a $3 \times 3 \times 3$ space, a solid $2 \times 2 \times 2$ cube would have a block count of 8 and negative space count of 0. However, a shape with only two blocks at opposite corners of the $2 \times 2 \times 2$ shape would have a negative space count of 6. Importantly, negative space is not simply the number of air blocks within the genome, but the number of air blocks within the minimal cuboid that bounds the shape. Some combinations of block count and negative space are impossible to achieve, i.e.\ a block count of 26 can only be paired with a negative space score of 1. Referred to as {\tt ME.CN} in results.



\item {\bf Piston Orientation:} Flying machines depend on having pistons that face in opposite directions so that they can push the shape back and forth. Therefore, categorizing shapes based on the count of pistons aimed in each direction makes sense. The pistons and sticky pistons in each shape are grouped into three categories based on their orientations: north or south, east or west, and up or down. Each of the three dimensions distinguishes between counts with 0, 1, 2, 3, 4, 5 or more pistons. So 6, 7, 8, etc. can occur but are characterized the same. Referred to as {\tt ME.PO} in results.

\end{enumerate}

\subsection{Pure Fitness-based Evolution}


To demonstrate the benefits of using a quality diversity algorithm, MAP-Elites is compared with a pure fitness-based approach. This approach demonstrates why flying machines are hard to evolve. However, in contrast to previous work in \emph{EvoCraft} \cite{grbic2021evocraft}, the pure fitness approach in this paper is occasionally successful. 

In the \emph{EvoCraft} \cite{grbic2021evocraft} paper, a simple evolutionary strategy (ES) optimizer was applied with a population size of 10. Specifically, the ES used was based on the work of Salimans \el~\cite{salimans:arxiv2017es}, and involves sampling genomes from a distribution that is updated based on the fitness scores of the genomes sampled at each generation.

The approach used in this paper is a more traditional evolution approach that maintains an actual population, upon which $(\mu + \lambda)$ selection is applied \cite{beyer:natcomS02}.
The population of $\mu$ parents is evaluated via binary tournament selection to create $\lambda$ children via both crossover and mutation, and the best $\mu$ members of the combined $(\mu + \lambda)$ population are selected to be the next generation via pure elitist selection.
Most ES algorithms do not incorporate crossover, but it is applied in this paper to maintain consistency with MAP-Elites. This approach is referred to as {\tt PF} in the results.

\subsection{Experimental Parameters}


Each experiment was run 30 times using each MAP-Elites behavior characterization and pure fitness-based evolution. Experiments were run with both the original block set and the observer block set. In MAP-Elites runs, 100 shapes were initially generated, followed by 60,000 more shapes, which provided a reasonable trade-off in terms of the time needed to produce flying machines vs.\ the time cost of running longer experiments. Parameters for fitness-based evolution were selected to evaluate the same number of individuals by setting $\mu = \lambda = 20$ and evolving for 3005 generations ($20  \times  3005$ = 60,100). 
Each genome had a 50\% crossover rate. During mutation, every individual index in the genome has an individual 30\% chance of being modified via polynomial mutation~\cite{deb1:cs95:polynomial}. 


\emph{Minecraft} itself is single-threaded, but the experiment code is multi-threaded. Each of 10 separate threads spawns a shape in the server, and \emph{Minecraft} simulates them and determines fitness as described in Section \ref{sec:fitness}. 
Shapes are placed diagonally along all three dimensions because they 
only fly in orthogonal directions. The diagonal placement prevents shapes that fly out of their starting area from crashing into each other. 
Shapes are generated in a $3 \times 3 \times 3$ space of 27 blocks, so each genome length is 81 ($3 \times 27$) values.
However, to assure that any displaced blocks from the previously evaluated shape will be cleared away, a $43 \times 43 \times 43$ space is cleared before each evaluation. This provides a buffer of 20 blocks on each side of the shape.
However, to register as a flying machine, the shape only needs to vacate a $9 \times 9 \times 9$ area centered on the shape. If over 6 blocks leave this space, then the shape is considered a flying machine. The $F_{\max}$ fitness value assigned is 55 before penalties for leftover blocks. This value is meant to exceed what would be awarded in the impossible scenario of the shape moving from one end of the evaluation area to the other multiple times within the evaluation time. Actual fitness scores for shapes that do not fly are less than 10, no matter how much they oscillate.




\section{Results}
\label{sec:results}

Quantitative and qualitative results are described below. The quantitative results compare the different behavior characterizations to each other and to pure fitness-based evolution. The qualitative results describe observations of evolved flying machines. The videos mentioned and other resources are all available online at \url{https://people.southwestern.edu/~schrum2/SCOPE/minecraft.php}

\subsection{Quantitative Results}

\begin{figure}[t]
\centering
\makebox[\columnwidth]{
\subfloat[Original Block Set]{
  \label{fig:originalResults}
  \makebox[0.5\columnwidth]{
  \includegraphics[width=0.5\columnwidth]{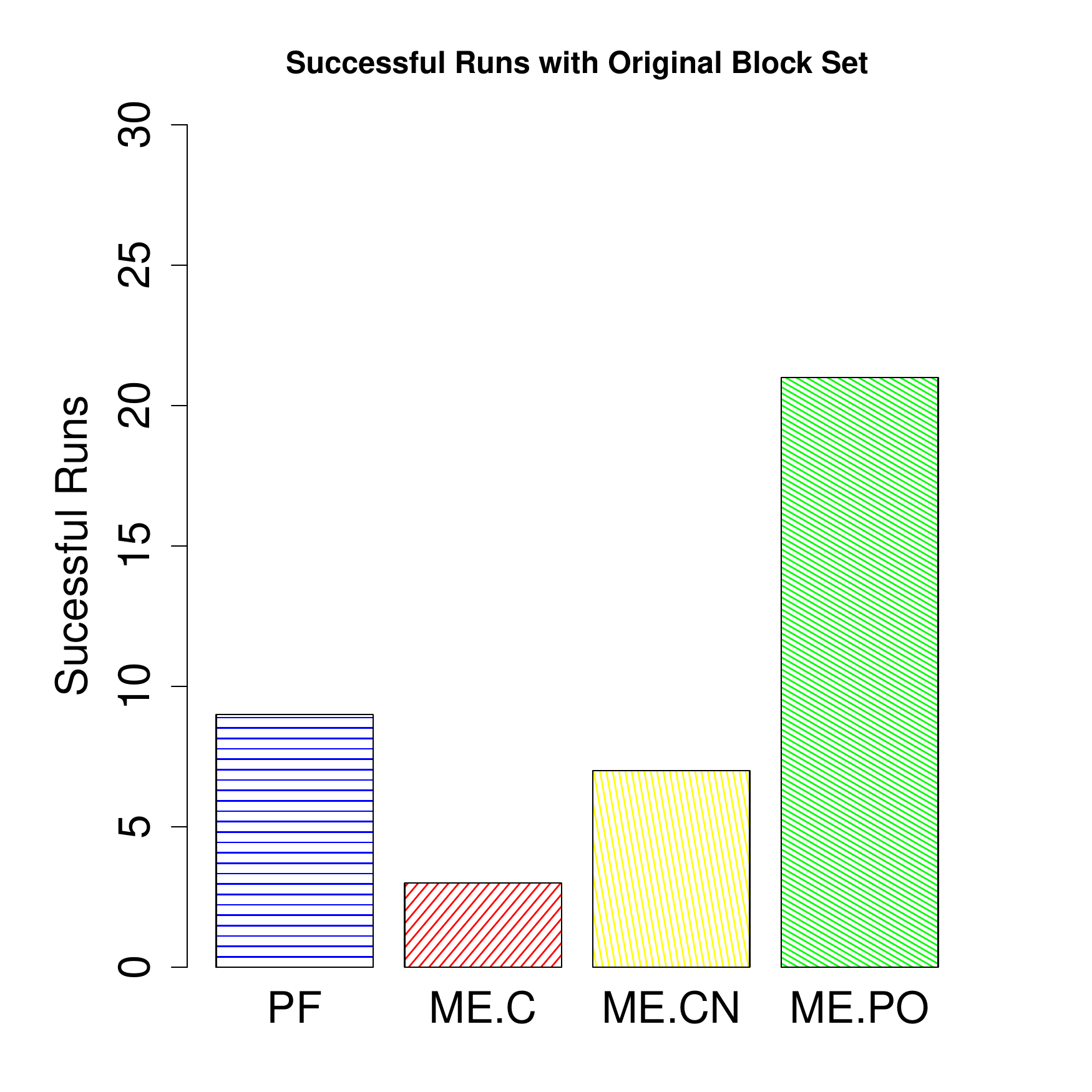}}} 
\subfloat[Observer Block Set]{
  \label{fig:observerResults}
  \includegraphics[width=0.5\columnwidth]{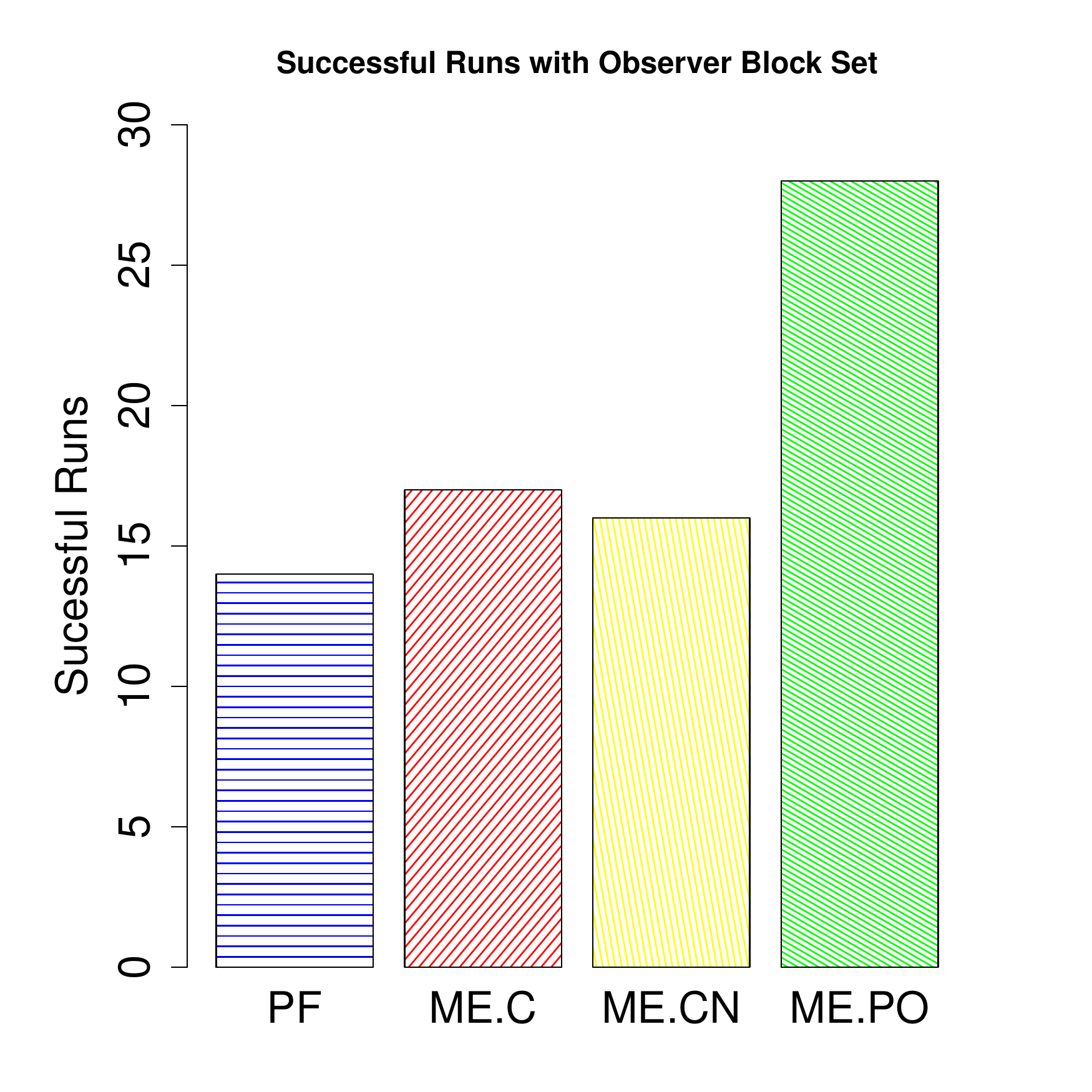}} 
}
\caption{Original and Observer Block Set Success Counts: (\protect\subref{fig:originalResults}) Number of runs producing flying machines using different behavior characterizations and pure fitness for the original block set. The pure fitness strategy was not the worst performing approach, but it is significantly less successful than Piston Orientation MAP-Elites. (\protect\subref{fig:observerResults}) Number of runs producing flying machines using different behavior characterizations and pure fitness for the observer block set. This figure also shows that Piston Orientation MAP-Elites was the most successful in generating flying machines. Pure fitness was comparable to but less successful than {\tt ME.C} and {\tt ME.CN}.}
\label{fig:successRate}
\end{figure}

The number of runs that discovered at least one flying machine are in both
Fig.~\ref{fig:successRate} and Table~\ref{tab:successRateTable}. Runs with the observer block set were more successful at producing flying machines than the original block set. Runs using Piston Orientation were the most successful with both block sets. When using the observer block set, pure fitness was slightly less successful than {\tt ME.C} and {\tt ME.CN}. However, in runs with the original block set, pure fitness had more successful runs than these two approaches.

\begin{table}[t]
\caption{\label{tab:successRateTable} Original and Observer Successful Runs}
Number of runs of each method that produced flying machines out of 30, along with associated percentages.
\begin{tabular}{|l|l|l|}
\hline
               & \textbf{Original} & \textbf{Observer} \\ \hline
\texttt{PF}    & 9 (33\%)          & 14 (46.67\%)      \\ \hline
\texttt{ME.C}  & 3 (10\%)          & 17 (56.67\%)      \\ \hline
\texttt{ME.CN} & 7 (23.33\%)       & 16 (53.33\%)      \\ \hline
\texttt{ME.PO} & 21 (70\%)         & 28 (93.33\%)      \\ \hline
\end{tabular}
\end{table}


%
%
%

%
%
%

Success rates were compared using pairwise Fisher exact tests with Bonferroni error correction to account for the increased risk of a Type II error from multiple comparisons. For the observer block set, {\tt PF}, {\tt ME.C}, and {\tt ME.CN} did not have significantly different success rates ($p \approx 1.0$). However, the success rate for {\tt ME.PO} is significantly higher than the success rates for {\tt PF} ($p \approx 0.00087$), {\tt ME.C} ($p \approx 0.01277$), and {\tt ME.CN} ($p \approx 0.00546$). Similarly, the original block set did not have significantly different success rates for {\tt PF}, {\tt ME.C}, and {\tt ME.CN}. The success rate for {\tt ME.PO} is significantly higher than the success rates for {\tt PF} ($p \approx 0.0247$), {\tt ME.C} ($p \approx 0.00002$), and {\tt ME.CN} ($p \approx 0.0038$). So, {\tt ME.PO} had significantly higher success rates for both block sets.

\begin{figure}[t]
\centering
\makebox[\columnwidth]{
\subfloat[Original Block Set]{
  \label{fig:originalResultsDirections}
  \makebox[0.5\columnwidth]{
  \includegraphics[width=0.5\columnwidth]{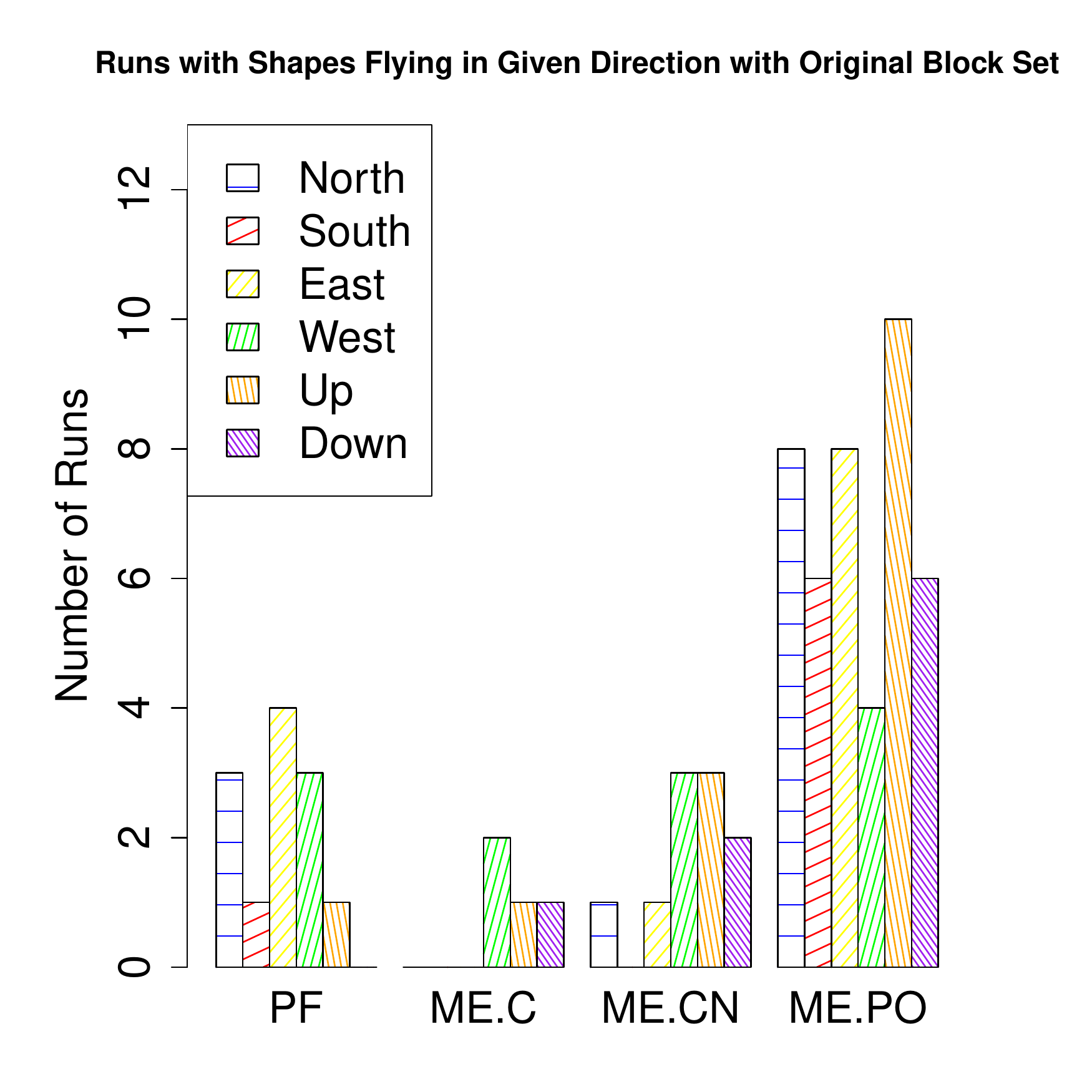}}} 
\subfloat[Observer Block Set]{
  \label{fig:observerResultsDirections}
  \includegraphics[width=0.5\columnwidth]{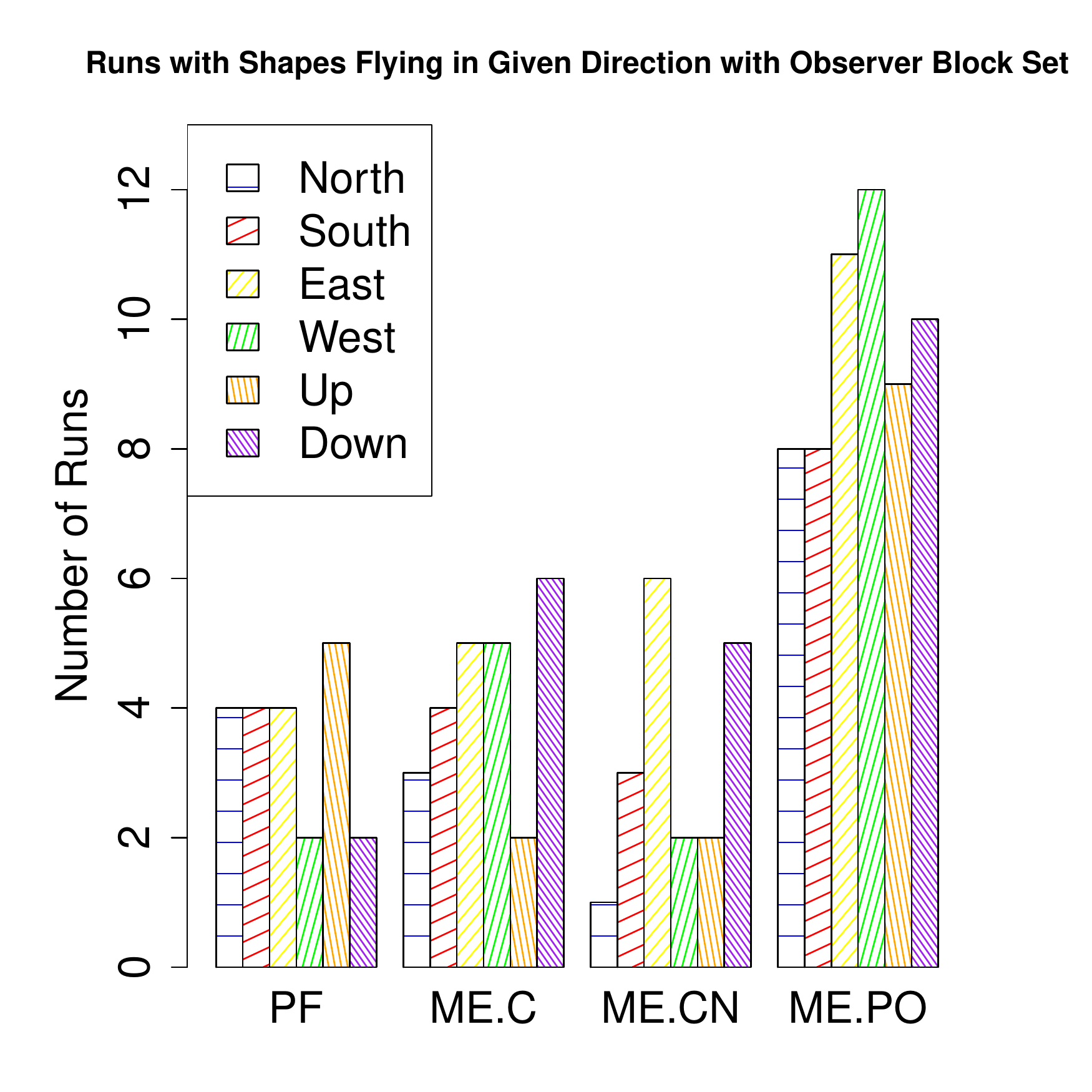}} 
}
\caption{Original and Observer Block Set Direction Counts: (\protect\subref{fig:originalResultsDirections}) Number of runs producing flying machines in each direction for pure fitness and MAP-Elites using the original block set. The most successful MAP-Elites approach was Piston Orientation, and it was the only approach that produced flying machines in every direction. (\protect\subref{fig:observerResultsDirections}) Number of runs producing flying machines in each direction for pure fitness and MAP-Elites using the observer block set. Piston Orientation was once again the most successful in generating flying machines for all six directions.}
\label{fig:successRateDirections}
\end{figure}

Fig.~\ref{fig:successRateDirections} shows how successful each approach was at generating shapes that fly in different directions. For each possible direction a machine could fly in, {\tt ME.PO} had several runs with machines flying in that direction. All other approaches had fewer runs with machines flying in each direction, and with the original block set, {\tt ME.PO} was the only approach with runs producing a machine for each direction. {\tt PF} and {\tt ME.CN} only produced machines flying in five directions, and {\tt ME.C} only produced machines in three of the six possible directions.
Results for the original block set are presented with more detail in Table~\ref{tab:originalResultsDirections}, and detailed observer block set results are in Table \ref{tab:observerResultsDirections}. These tables also show the average and maximum number of distinct flight directions discovered across individual runs of each type, and show that {\tt ME.PO} not only discovered more directions on average, but was the only approach with runs where shapes flew in four different directions. {\tt ME.C} also had one run with shapes flying in three directions using the observer block set, but all other methods maxed out at two directions.

\begin{table}[t]
\caption{\label{tab:originalResultsDirections} Directions Discovered Using Original Block Set}
The number of runs of each method using the original block set that produced any flying machines in each specific direction out of 30, along with associated percentages. The last two rows are the average and maximum number of distinct directions that machines within individual runs flew in.
\begin{tabular}{|l|l|l|l|l|}
\hline
\textbf{} & \texttt{PF} & \texttt{ME.C} & \texttt{ME.CN} & \texttt{ME.PO} \\ \hline
\textbf{North}    & 3 (10\%)    & 0 (0\%)       & 1 (3.33\%)     & 8 (26.67\%)    \\ \hline
\textbf{South}    & 1 (3.33\%)  & 0 (0\%)       & 0 (0\%)        & 6 (20\%)       \\ \hline
\textbf{East}     & 4 (13.33\%) & 0 (0\%)       & 1 (3.33\%)     & 8 (26.67\%)    \\ \hline
\textbf{West}     & 3 (10\%)    & 2 (6.67\%)    & 3 (10\%)       & 4 (13.33\%)    \\ \hline
\textbf{Up}       & 1 (3.33\%)  & 1 (3.33\%)    & 3 (10\%)       & 10 (33.33\%)   \\ \hline
\textbf{Down}     & 0 (0\%)     & 1 (3.33\%)    & 2 (6.67\%)     & 6 (20\%)       \\ \hline
\textbf{Avg \#}   & 0.4         & 0.13        & 0.33          & 1.4            \\ \hline
\textbf{Max \#}   & 2           & 2             & 2              & 4              \\ \hline
\end{tabular}
\end{table}


\begin{table}[t]
\caption{\label{tab:observerResultsDirections} Directions Discovered Using Observer Block Set}
Results presented as in Table \ref{tab:originalResultsDirections}, but for the observer block set.
\begin{tabular}{|l|l|l|l|l|}
\hline
\textbf{} & \texttt{PF} & \texttt{ME.C} & \texttt{ME.CN} & \texttt{ME.PO} \\ \hline
\textbf{North}    & 4 (13.33\%) & 3 (10\%)      & 1 (3.33\%)     & 8 (26.67\%)    \\ \hline
\textbf{South}    & 4 (13.33\%) & 4 (13.33\%)   & 3 (10\%)       & 8 (26.67\%)    \\ \hline
\textbf{East}     & 4 (13.33\%) & 5 (16.67\%)   & 6 (20\%)       & 11 (36.67\%)   \\ \hline
\textbf{West}     & 2 (6.67\%)  & 5 (16.67\%)   & 2 (6.67\%)     & 12 (40\%)      \\ \hline
\textbf{Up}       & 5 (16.67\%) & 2 (6.67\%)    & 2 (6.67\%)     & 9 (30\%)       \\ \hline
\textbf{Down}     & 2 (6.67\%)  & 6 (20\%)      & 5 (16.67\%)    & 10 (33.33\%)   \\ \hline
\textbf{Avg \#}   & 0.7         & 0.87        & 0.63          & 1.93        \\ \hline
\textbf{Max \#}   & 2           & 3             & 2              & 4              \\ \hline
\end{tabular}
\end{table}

\begin{figure}[t]
\centering
\makebox[\columnwidth]{
\subfloat[Original Block Set]{
  \label{fig:originalTimeToSuccess}
  \makebox[0.47\columnwidth]{
  \includegraphics[width=0.47\columnwidth]{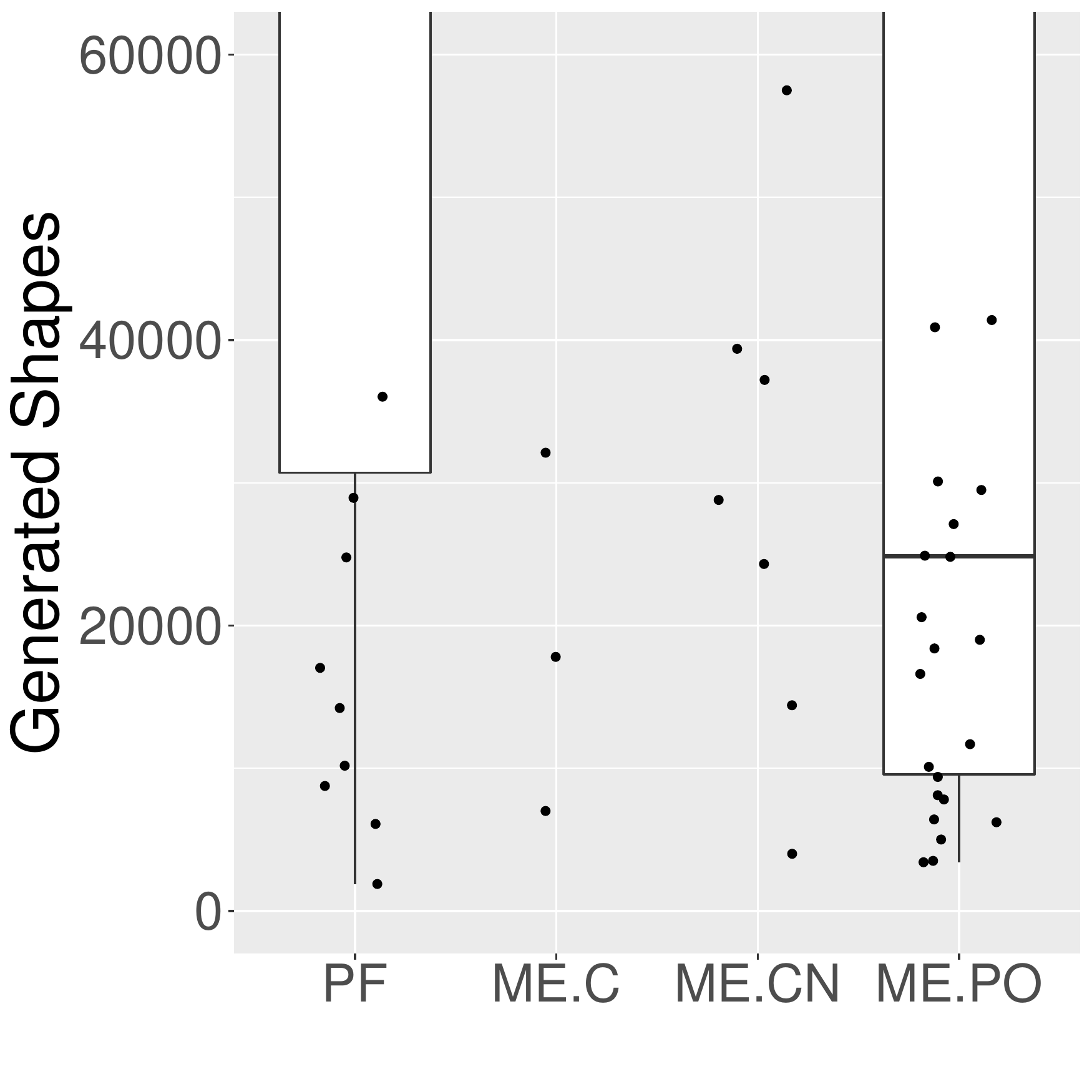}}} 
\subfloat[Observer Block Set]{
  \label{fig:observerTimeToSuccess}
  \includegraphics[width=0.47\columnwidth]{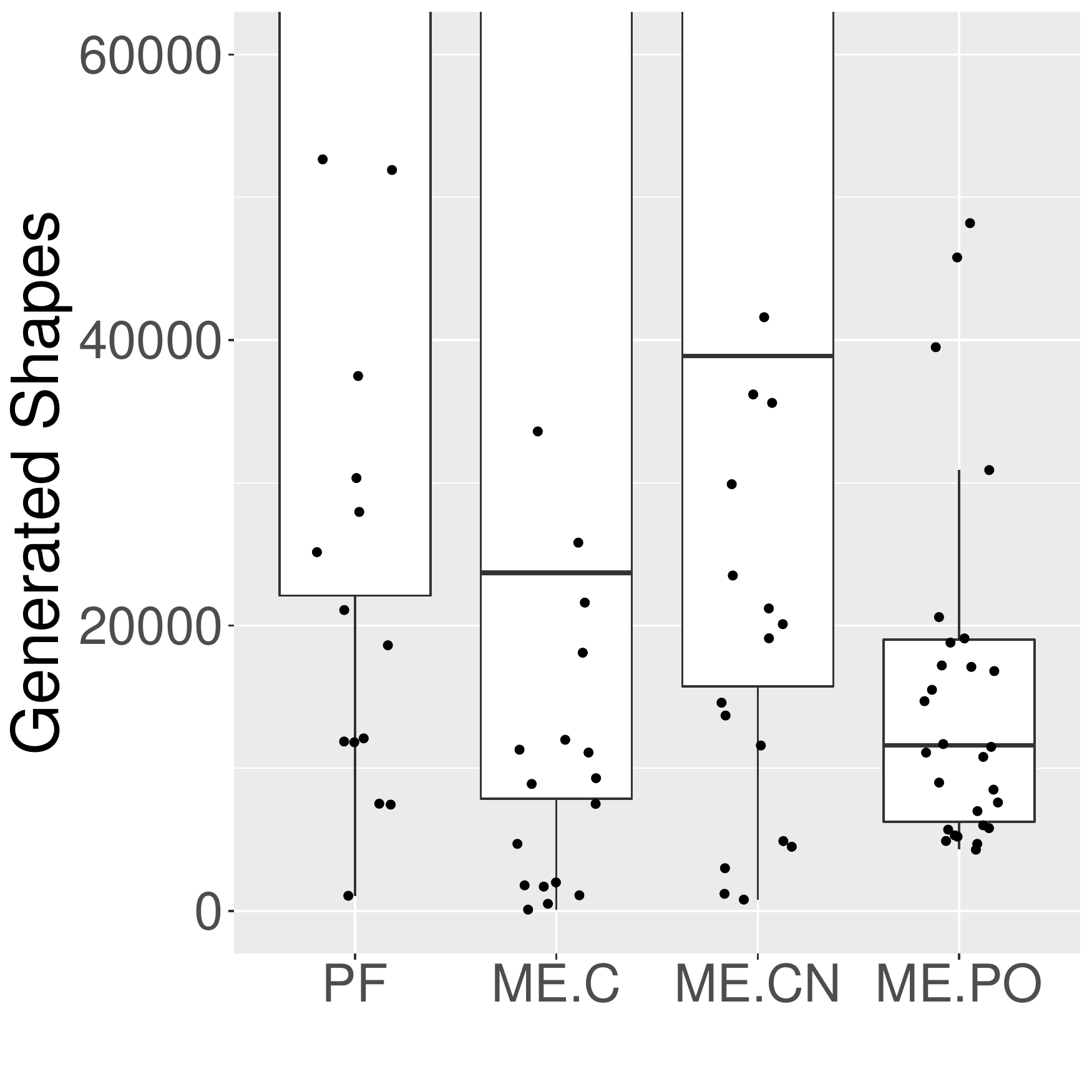}} 
}
\caption{Original and Observer Block Set Number Of Generated Shapes Before First Flying Machine. Results are presented using box-and-whisker plots, which normally depict the minimum, lower quartile, median, upper quartile, and maximum of a distribution. However, runs that never generated a successful flying machine have an effective value of infinity, which is off the edge of the plot. Individual data points are shown with horizontal jitter to prevent overlapping.
(\protect\subref{fig:originalTimeToSuccess}) For the original block set, the minimum and lower quartile are visible for {\tt PF}, but over half of the runs failed, so the median is not visible. {\tt ME.C} and {\tt ME.CN} are even worse. Because over 75\% of their runs failed, only outlier points are shown. {\tt ME.PO} is the only approach whose median is visible, since it had many successful runs, which produced flying machines more quickly than other approaches. 
(\protect\subref{fig:observerTimeToSuccess}) For the observer block set, the median for {\tt PF} is still not visible, but the medians for all MAP-Elites approaches are visible. In fact, the two failed runs of {\tt ME.PO} (not visible) are outliers, as the distribution for {\tt ME.PO} is compact. However, the lowest score for {\tt ME.PO} is higher than the lowest of all other approaches. 
The upper quartiles for {\tt ME.C} and {\tt ME.CN} are still not visible.}
\label{fig:TimeToSuccess}
\end{figure}

For each approach, the distributions of the number of individuals generated before discovery of the first successful flying machine are shown in Fig.~\ref{fig:TimeToSuccess}. For approaches where less than a quarter of all runs succeeded, the few successes are treated as outliers. For the original block set, only {\tt ME.PO} has a visible median, but with the observer block set, the only approach whose median is not visible is {\tt PF}. Since the median is defined for all MAP-Elites approaches when using the observer block set, these results can be compared with Mood's Median Test. Note that data was only logged after each 100 individuals generated, so all reported medians are rounded up to the nearest 100 after when the first flying machine was actually generated. Significant differences between the median scores of {\tt ME.C}, {\tt ME.CN}, and {\tt ME.PO} were found ($\chi^{2} = 13.42885, p \approx 0.001213281$). However, follow-up pairwise testing indicates specifically that the median of 11,600 for {\tt ME.PO} is significantly less than the median 38,900 of {\tt ME.CN} ($p < 0.05$), but the median of 23,700 for {\tt ME.C} is not significantly different from the other two. 

These results all indicate superior performance by {\tt ME.PO} in terms of success rate, diversity of directions flown in, and time to success. However, further insight into differences between methods comes from actually looking at the evolved flying machines.

%
%
%
%
%
%

\subsection{Qualitative Results}


\begin{figure*}[h]

\makebox[\textwidth]{
\subfloat[Start]{
  \label{fig:start}
  \includegraphics[width=0.15\textwidth]{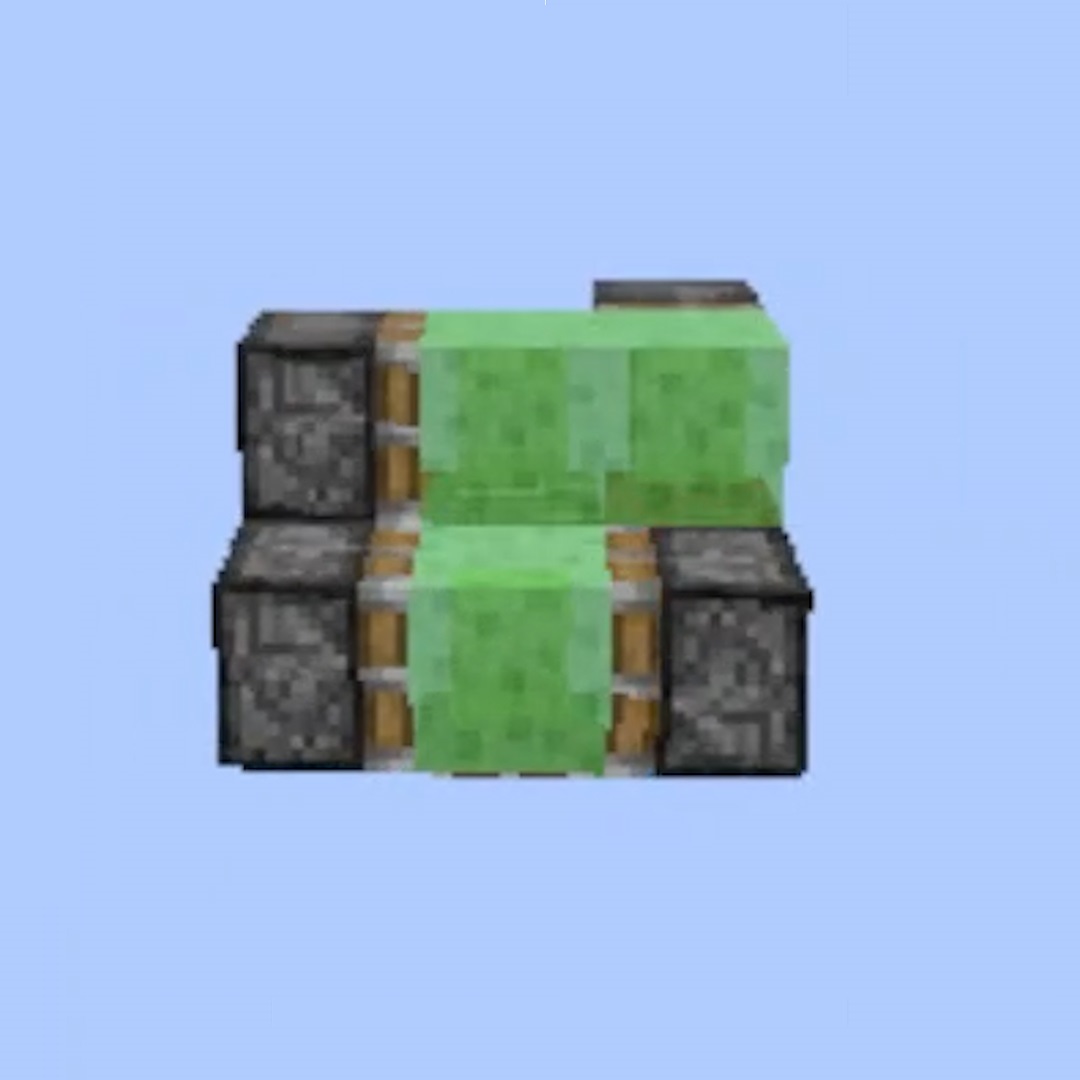}}%
\subfloat[Push Right]{
  \label{fig:right1}
  \includegraphics[width=0.15\textwidth]{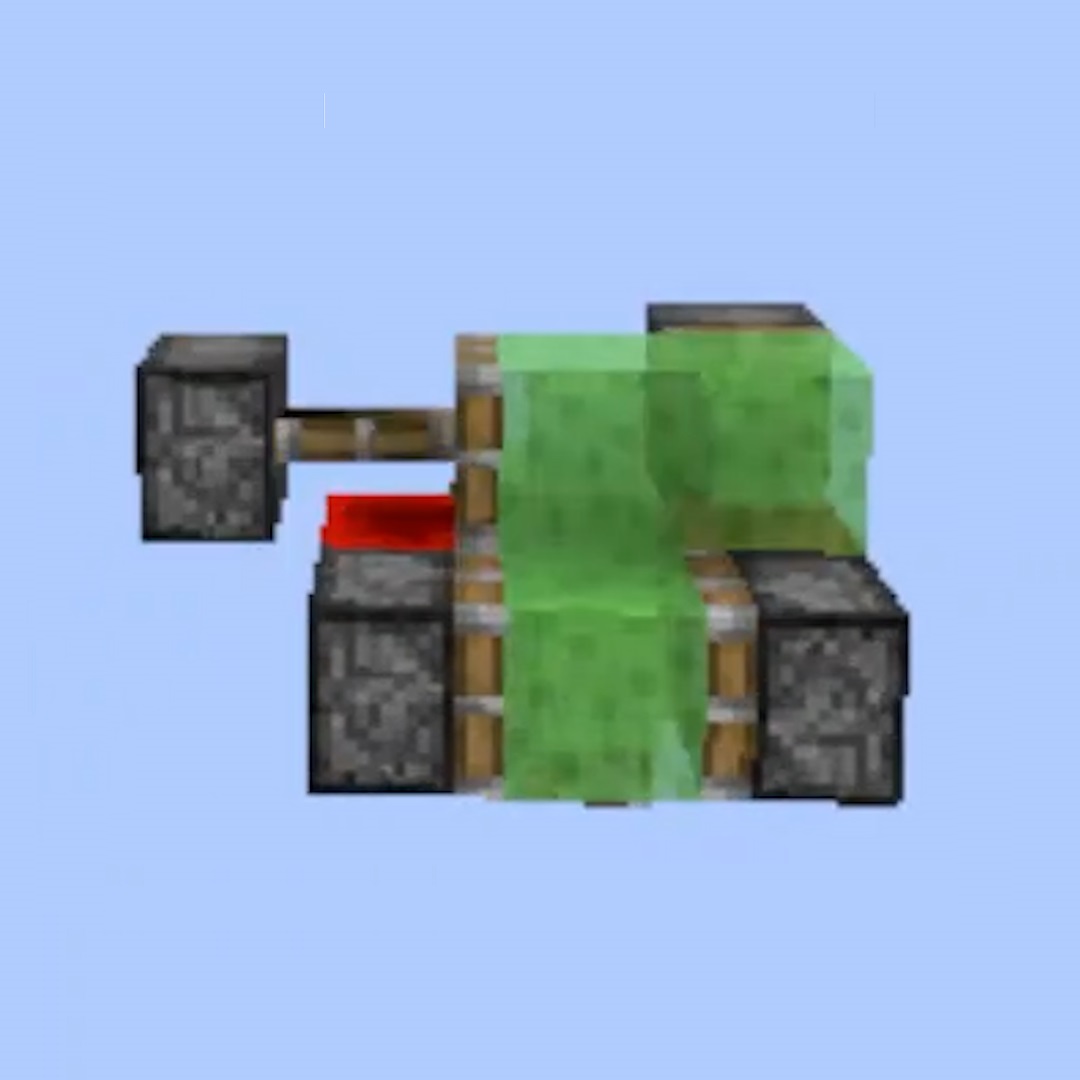}}%
\subfloat[Push Right]{
  \label{fig:right2}
  \includegraphics[width=0.15\textwidth]{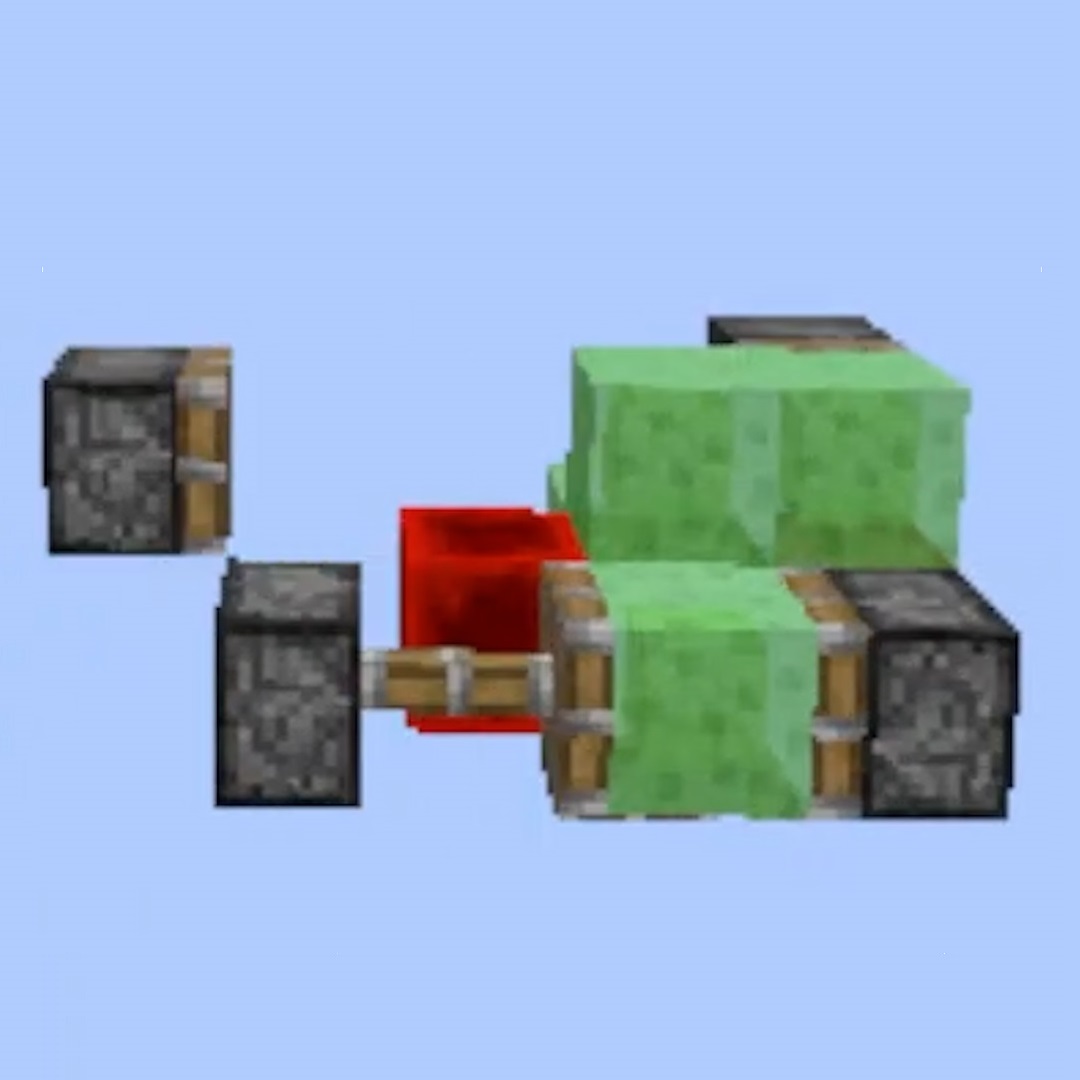}}%
\subfloat[Retract Left]{
  \label{fig:retractLeft}
  \includegraphics[width=0.15\textwidth]{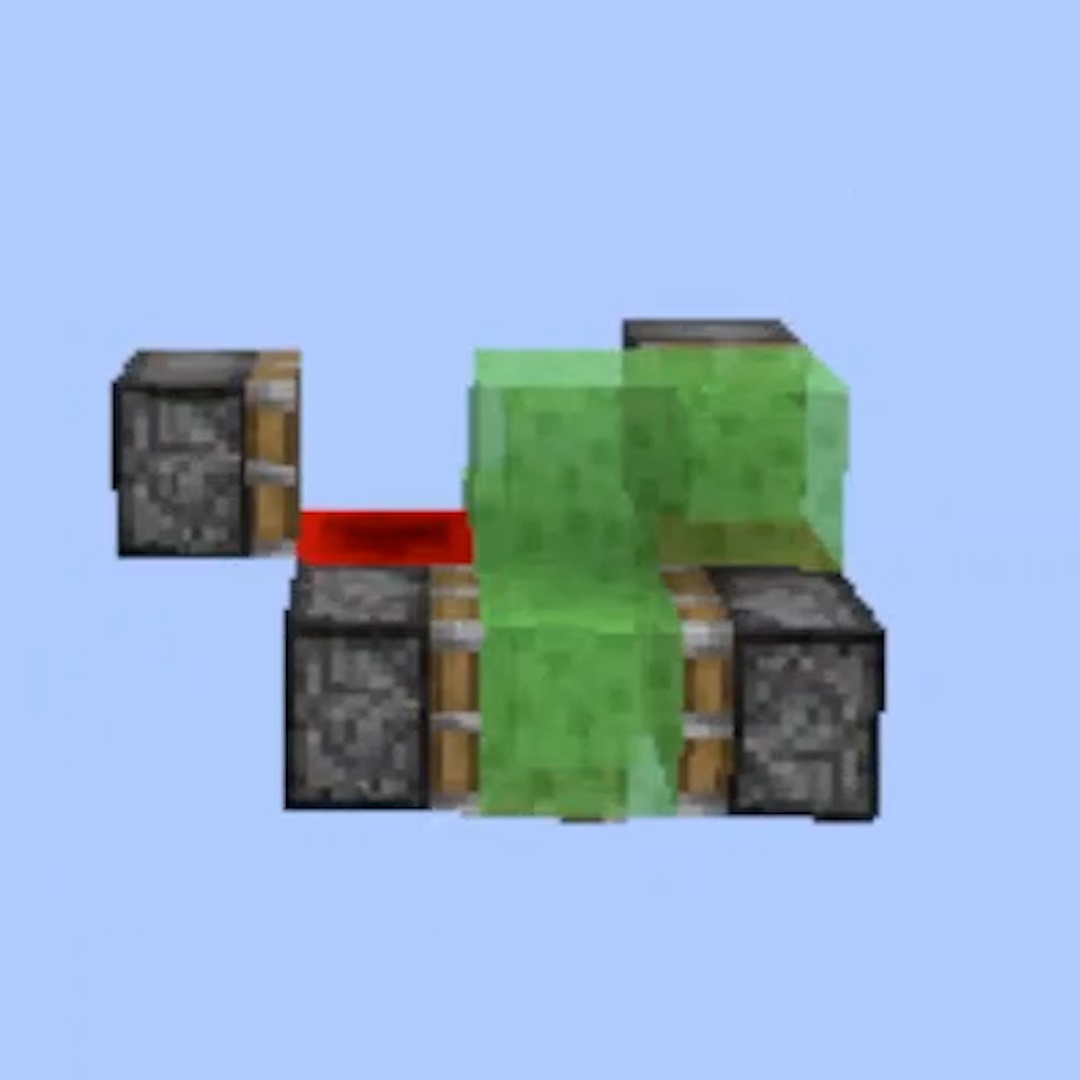}}%
\subfloat[Push Left]{
  \label{fig:left}
  \includegraphics[width=0.15\textwidth]{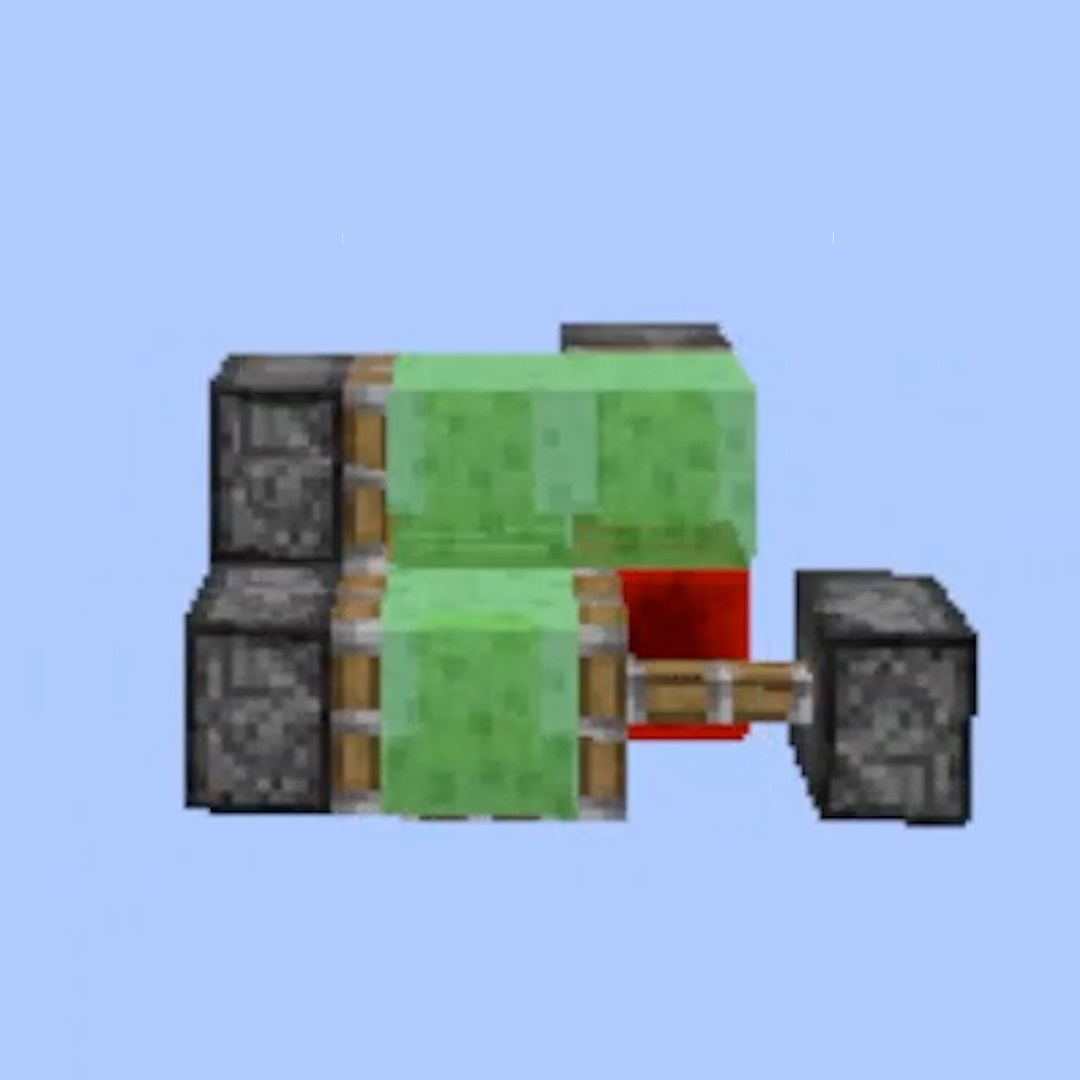}}%
\subfloat[Retract Right]{
  \label{fig:rightRetract}
  \includegraphics[width=0.15\textwidth]{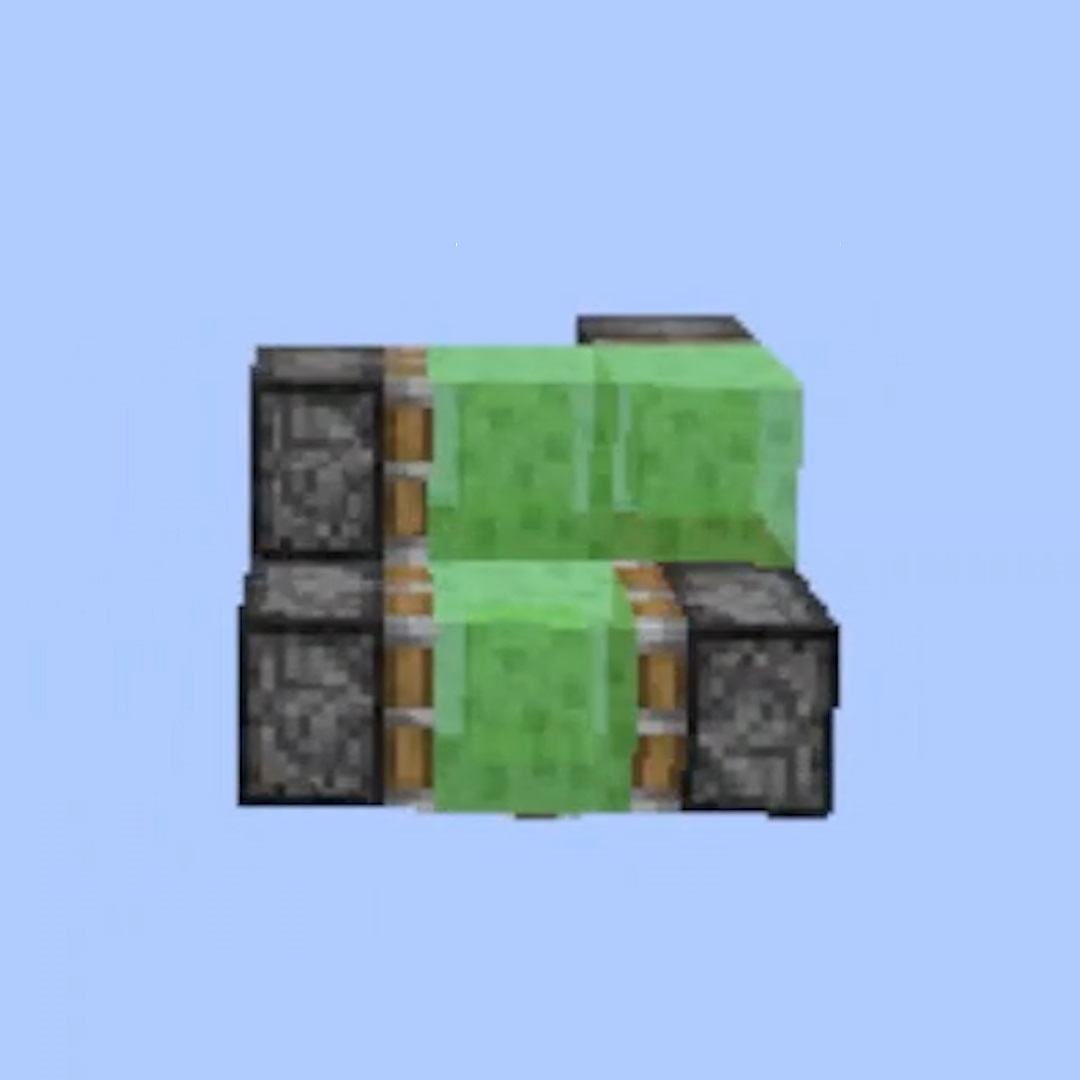}}%
}
\caption{A Typical Flying Machine. From (\subref{fig:start}), two different pistons push the shape to the right in (\subref{fig:right1}) and (\subref{fig:right2}), but after retracting in (\subref{fig:retractLeft}), a sticky piston pushes to the left in (\subref{fig:left}) where slime blocks attach to the pistons that were pushing to the right. As a result, when the leftward facing sticky piston in (\subref{fig:rightRetract}) retracts, it pulls the whole shape one block to the right before restarting the cycle. Video of this behavior is at \url{https://tinyurl.com/FlyToRight}}
\label{fig:exampleMachine}
\end{figure*}

\begin{figure*}[h]

\makebox[\textwidth]{
\subfloat[Up]{
  \label{fig:leftmostMEC}
  \includegraphics[width=0.115\textwidth]{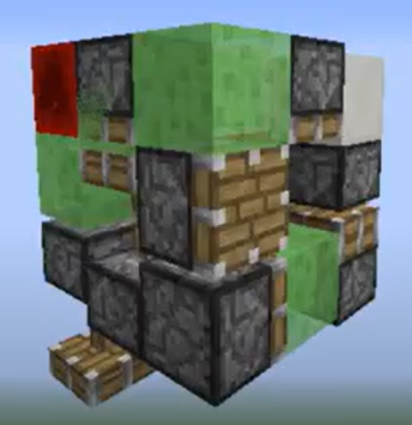}}%
\subfloat[Up]{
  \includegraphics[width=0.115\textwidth]{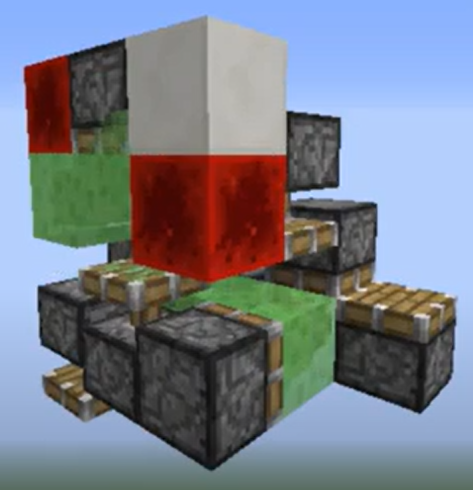}}%
\subfloat[Up]{
  \includegraphics[width=0.115\textwidth]{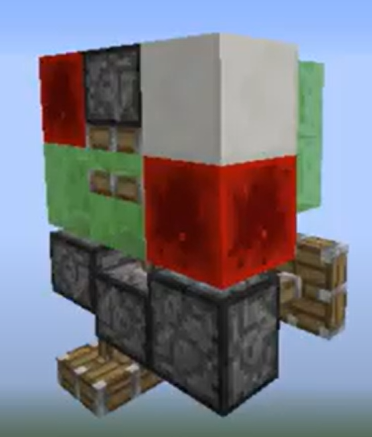}}%
\subfloat[Up]{
  \includegraphics[width=0.115\textwidth]{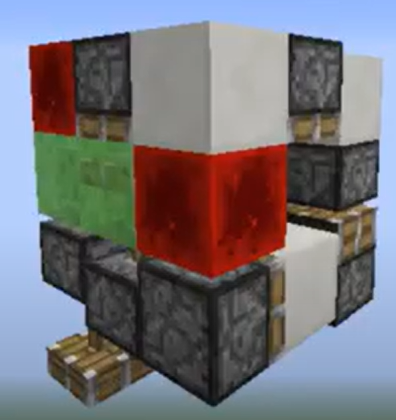}}%
\subfloat[Down]{
  \includegraphics[width=0.115\textwidth]{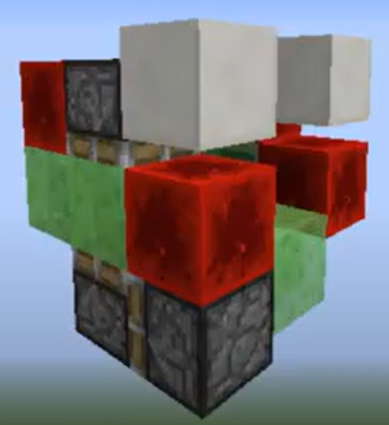}}%
\subfloat[Down]{
  \includegraphics[width=0.115\textwidth]{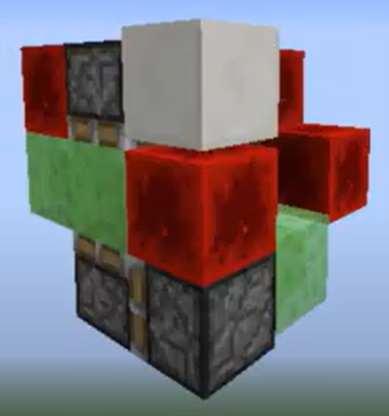}}%
  \subfloat[Down]{
  \includegraphics[width=0.115\textwidth]{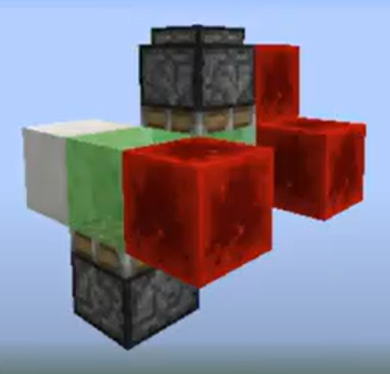}}%
  \subfloat[Down]{
  \includegraphics[width=0.115\textwidth]{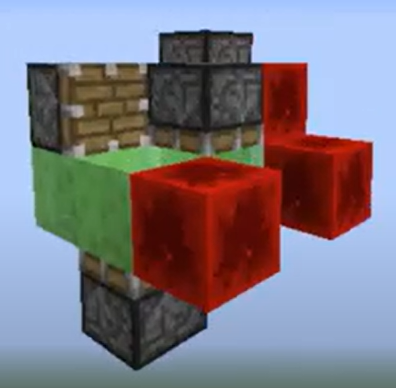}} 
}
\caption{Flying Machines From One Archive Evolved With Block Count Behavior Characterization and the Original Block Set. Each machine flies either up or down as indicated by its sub-caption, but all are similar. All but the leftmost machine in (\subref{fig:leftmostMEC}) have a redstone block in the center with a slime block to the left. All of them have an upward facing piston in the bottom-left center slot, regardless of whether they move up or down. Machines that move in the same direction move in the same way, but with different configurations of superfluous blocks. Machines that fly up or down
are using the same core pistons for movement, 
but the presence of additional pistons can disrupt the firing pattern and reverse the direction of movement. Video of these machines flying can be seen at \url{https://tinyurl.com/OriginalME_10}}
\label{fig:archiveMEC10}
\end{figure*}

\begin{figure*}[h]

\makebox[\textwidth]{
\subfloat[South]{
  \includegraphics[width=0.115\textwidth]{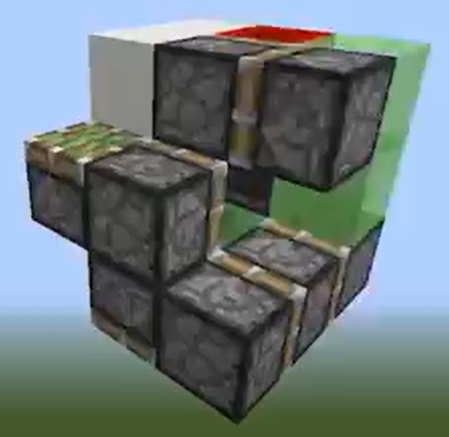}}%
\subfloat[South]{
  \includegraphics[width=0.115\textwidth]{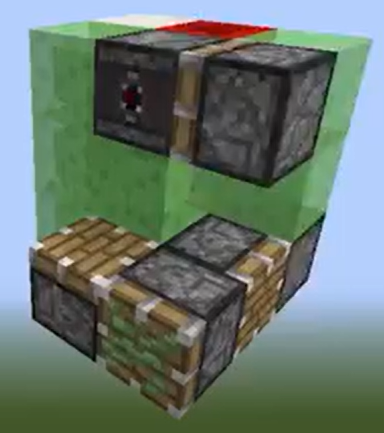}}%
\subfloat[South]{
  \includegraphics[width=0.115\textwidth]{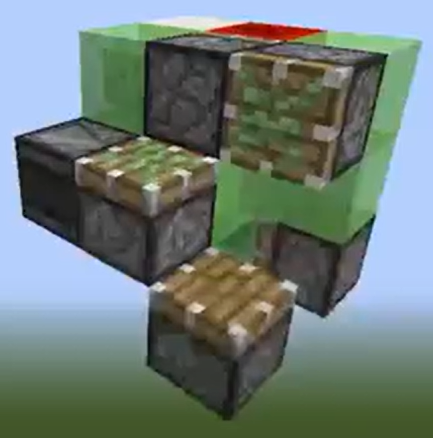}}%
\subfloat[West]{
  \label{fig:leftmostWestMEPO}
  \includegraphics[width=0.115\textwidth]{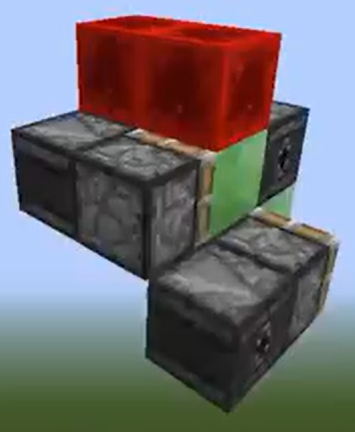}}%
\subfloat[West]{
  \includegraphics[width=0.115\textwidth]{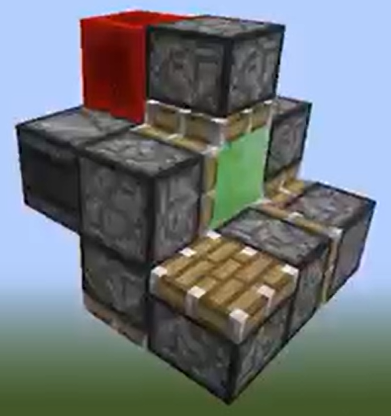}}%
\subfloat[West]{
  \includegraphics[width=0.115\textwidth]{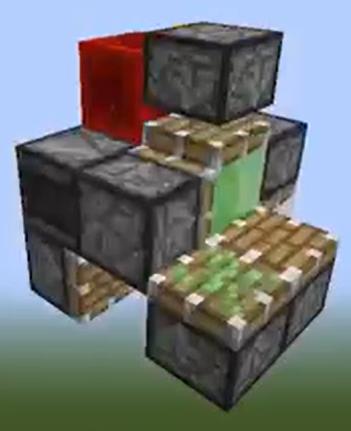}}%
  \subfloat[East]{
  \label{fig:eastMEPO}
  \includegraphics[width=0.115\textwidth]{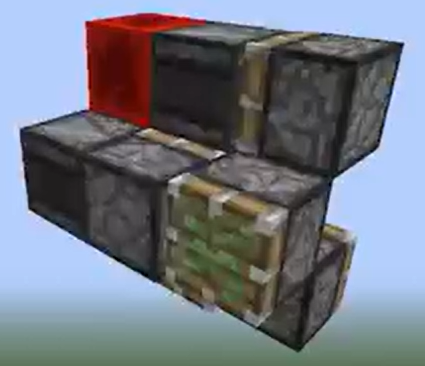}}%
  \subfloat[Down]{
  \label{fig:downMEPO}
  \includegraphics[width=0.115\textwidth]{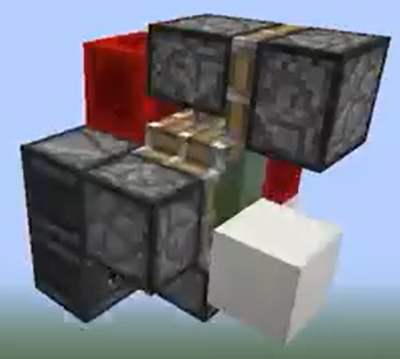}}%
}
\caption{Flying Machines From One Archive Evolved With Piston Orientation Behavior Characterization and the Observer Block Set. This archive features machines flying in four different directions, although there was only one machine that flew east and one that flew down. Most flew south or west. Machines that flew south had a similar set of pistons that determined movement, but some variation in superfluous blocks. Many machines that flew west started by being pushed down before moving west, but some flew directly westward. The leftmost westward flying machine in (\subref{fig:leftmostWestMEPO}) flies directly to the west, but the other two have a piston near the top that pushes the whole shape down first. The eastward flying machine in (\subref{fig:eastMEPO}) has a distinct structure, but the downward machine in (\subref{fig:downMEPO}) is similar to some of the westward ones, except it keeps moving down rather than changing direction after an initial push.
Video of these machines flying can be seen at \url{https://tinyurl.com/ObserverME_O06}}
\label{fig:archiveMEPO06}
\end{figure*}

Typical flying machine behavior is demonstrated in Fig.~\ref{fig:exampleMachine}. At a minimum, a flying machine needs to have pistons pointing in opposite directions, one of which should be a sticky piston. Blocks need to be configured so that the regular piston or pistons push part of the shape away, which then causes a sticky piston to reach back to grab the part left behind before retracting to pull it forward as well. There are many variations on this basic idea. Additional pistons could cause extra oscillations, resulting in one part of the shape moving back and forth twice for every one step that the whole shape moves forward. Sometimes movement orthogonal to the main direction of movement will also occur, but ultimately some form of back and forth piston pushing is required.



However, the variety of flying machines produced by each approach differs. {\tt PF}, {\tt ME.C}, and {\tt ME.CN} shapes generally had less variety. There was variation across runs, but
flying machines within a given run tended to all be similar. 
There are many runs where the only differences between flying machines were additional superfluous blocks that would move with the shape, or get left behind. Several flying machines from an {\tt ME.C} archive are shown in Fig.~\ref{fig:archiveMEC10} to demonstrate this point.
Despite additional blocks, the flying machines mostly flew in the same manner. 
Sometimes a rare shape would exhibit an unusual opening movement, for example moving to the side before beginning to move in the same direction and manner as other shapes produced within the run.



In contrast, {\tt ME.PO} produced more variation in flying machines both within runs and across runs. Flying machines in multiple directions were generated more frequently, so the way they flew also exhibited more variety, as seen in Fig.~\ref{fig:archiveMEPO06}. 
However, these shapes still had superfluous blocks like other approaches, and different shapes that moved in the same direction did tend to be similar.

Some shapes had very convoluted movement patterns. Such patterns were a bit more common with {\tt ME.PO} because the placement of pistons in a variety of different orientations was encouraged.
For example, some flying machines have an additional piston that pushes in a different direction than the rest of the shape is moving. This piston would not alter the movement of the shape, but would be part of the superfluous blocks that were activated as the rest of the shape moved. 
A few rare shapes exhibited a zig-zagging motion
involving alternating side to side movements,
even though the general direction of movement would be up or down. An example can be seen at \url{https://tinyurl.com/MEPOUpward}.

\section{Discussion and Future Work}
\label{sec:discussion}


MAP-Elites performed better than 
fitness-based evolution when using the Piston Orientation behavior characterization, but not always with others. Piston Orientation was also the best at producing machines that could fly in a variety of different directions. 
Block Count performed the worst in comparison to the other behavior characterizations, and was worse than pure fitness when using the original block set, although it was better with the observer block set. Most runs that did not use Piston Orientation only produced machines that flew in one or two directions. In fact, there is only a single exception: one {\tt ME.C} run produced machines flying in three directions when using the observer block set.


Piston Orientation encourages the placing of varying numbers of
pistons aimed along each axis. Successful flying machines have pistons that are oriented in opposing directions. Other behavior characterizations do not consider the number of pistons in a given machine, which results in lower success rates. It is hard to generalize the success of {\tt ME.PO} to other domains, though it does make sense in retrospect that experimenting with different piston placements is more relevant to the goal of discovering flying machines.



The reason {\tt PF} flying machines within a given run are so similar is that elitist selection is used. Once a flying machine is discovered, the only mutations that are allowed are those that do not meddle with the shape's ability to fly, and the whole population converges around essentially the same solution. Larger populations and a less harsh selection mechanism could have allowed for more variance, but there would still not be any explicit push for diversity, making local optima a persistent challenge to overcome.

Flying machines in {\tt ME.C} runs suffer from a similar lack of diversity.
In these runs, a basic arrangement of pistons that could fly would emerge, and this resulted in the evolution of other shapes that were nearly identical but for the exception of a few superfluous blocks being added or removed. 
The ease with which this trick allowed one flying machine to dominate a range of bins with its offspring stifled the discovery of diverse solutions.

{\tt ME.CN} runs seem to lack diversity for the same reason. Despite 
the extra dimension of Negative Space, this behavior characterization is still based on Block Count, and supplanting nearby bins by only adding or removing superfluous blocks still proved effective.

It makes sense for {\tt PF}, {\tt ME.C}, and {\tt ME.CN} to converge to one kind of flying machine since they do not consider the importance of diversity in flight direction. Modifying a machine so that it flies in a different direction requires the functional components to be changed, not just the superfluous blocks, which is why {\tt ME.PO} leads to more diverse behaviors.


With all methods, if a given run evolved shapes that flew in one direction, the next direction that was most likely to be discovered would be the opposite direction. 
A flying machine must already have pistons aimed in opposite directions along one axis, and often only a small adjustment is needed to a shape in order to make the existing pistons activate with a different timing sequence, which can reverse a shape's movement direction.

One may wonder if producing shapes that fly in a variety of different directions is actually impressive.
In particular, it would be reasonable to assume that any shape that flies in one direction could simply be rotated to fly in any other direction, but the mechanics of \emph{Minecraft} make this assumption oddly untrue. Some machines that fly in one direction will not fly if completely reoriented to aim in a different direction. Furthermore, the unusual bug in \emph{EvoCraft} that prevents observers from being oriented up or down (Section \ref{sec:genome}) means that certain machines cannot be faithfully instantiated in certain orientations, so discovering distinct solutions for different directions is more impressive.




The experiments in this paper apply MAP-Elites to \emph{Minecraft} with a particular fitness function and behavior characterizations. However, it is clear that changing the behavior characterization has a large impact on what is evolved, and changing the fitness function could have an even bigger impact.
Different fitness functions, block sets, and behavior characterizations should lead to a variety of evolved artefacts with varying functional and aesthetic properties.




\emph{Minecraft} also presents an opportunity to explore how human interaction can affect quality diversity search.
The archive of previously evolved shapes can be represented in the game world in a way that humans can interact with the shapes, and simple game mechanics can be used to allow a player to influence which shapes remain in the archive or get selected for reproduction. Exploring these possibilities is an interesting avenue of future research.

Ultimately, \emph{Minecraft} is just one example of a game where Procedural Content Generation has a significant impact on the user experience. Rich and varied landscapes are already procedurally generated within the game, but procedurally generating functioning machines could add an interesting dimension to the game. Such techniques for content generation are relevant for other games too.

\section{Conclusion}

MAP-Elites has demonstrated an ability to reliably generate flying machines in \emph{Minecraft}. It is  possible to evolve flying machines using only fitness, but quality diversity was more successful with the right behavior characterization. This paper presents the first method for using evolution to generate flying machines in \emph{Minecraft}. \emph{Minecraft} is an open-ended domain worth further exploration, and there are still many interesting artefacts that can be evolved in the future. 

\begin{acks}

This research was made possible by the donation-funded Summer Collaborative Opportunities and Experiences (SCOPE) program for undergraduate research at Southwestern University

\end{acks}

\bibliographystyle{ACM-Reference-Format}


\end{document}